\documentclass{article}
\usepackage{stywhispers,amsmath,epsfig}

\usepackage{microtype}
\usepackage{graphicx}
\usepackage{subfigure}
\usepackage{booktabs} 

\usepackage{hyperref}

\usepackage{amsmath}
\usepackage{amssymb}
\usepackage{mathtools}
\usepackage{amsthm}
\usepackage{qtree}

\usepackage[capitalize,noabbrev]{cleveref}

\theoremstyle{plain}

\theoremstyle{definition}

\theoremstyle{remark}

\title{Classifying Crop Types using Gaussian Bayesian Models and Neural Networks on GHISACONUS USGS data from NASA Hyperspectral Satellite Imagery}

\name{Bill Basener}
\address{University of Virginia, School of Data Science \\
Charlottesville, VA \\
wb8by@virginia.edu}

\begin{document}
%
\maketitle
\begin{abstract}
Hyperspectral Imagining is a type of digital imaging in which each pixel contains typically hundreds of wavelengths of light providing spectroscopic information about the materials present in the pixel~\cite{Olsen_Bergman_Resmini_1997}.  In this paper we provide classification methods for determining crop type in the USGS GHISACONUS data, which contains around 7,000 pixel spectra from the five major U.S. agricultural crops (winter wheat, rice, corn, soybeans, and cotton) collected by the NASA Hyperion satellite, and includes the spectrum, geolocation, crop type, and stage of growth for each pixel.  We apply standard LDA and QDA as well as Bayesian custom versions that compute the joint probability of crop type and stage, and then the marginal probability for crop type, outperforming the non-Bayesian methods.  We also test a single layer neural network with dropout on the data, which performs comparable to LDA and QDA but not as well as the Bayesian methods.
\end{abstract}

\begin{keywords}
Machine Learning, hyperspectral, USGS, GHISACONUS, crop identification, vegetation classification, rice, wheat, corn, cotton, soybean, Bayesian
\end{keywords}

\section{Introduction}
Hyperspectral imaging is a digital imaging technology in which each pixel contains not just the usual three visual colors (red, green, blue) but many, often hundreds, of wavelengths of light enabling spectroscopic information about the materials located in the pixel~\cite{SchottBook2007}.  A common range of wavelengths covers the VNIRSWIR (visible, near-infrared, short-wave infrared) from about 450nm to 2500nm, where the visible red, green, and blue colors correspond approximately to 650nm, 550nm, and 450nm respectively.  In each wavelength, some percentage of light is absorbed by the material on the ground and some percentage is reflected back, so that the measured spectrum is a vector of percent reflectance for each wavelength.  The absorption or reflection depends mainly on the interaction of the wavelength of the light and molecular elements and bonds present, enabling some level spectroscopy - determining information about the materials present from the measured spectrum

\subsection{Data}
\begin{figure}[ht]
\vskip 0.2in
\begin{center}
\centerline{\includegraphics[width=\columnwidth]{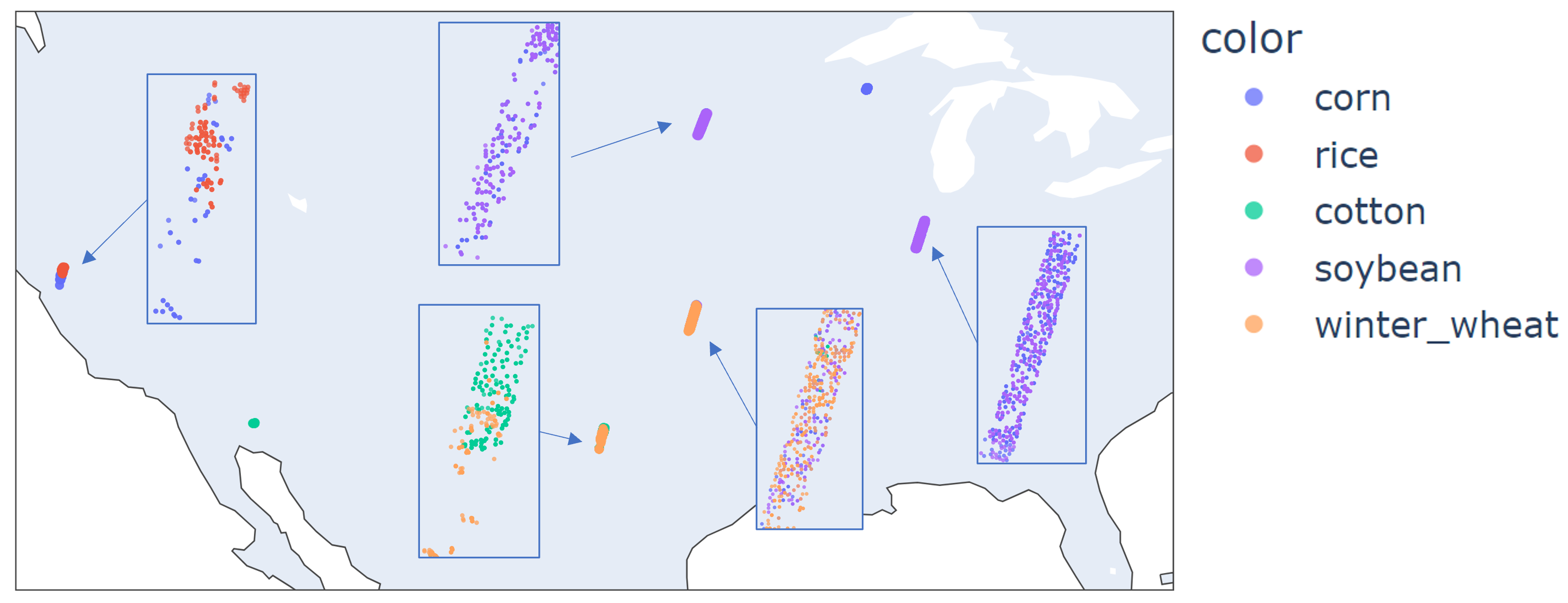}}
\caption{Locations of the crops associated with the collected spectra, with each spectrum indicated by a marker colored by the crop type.  Observe that the collections come in rectangular regions, each of which corresponds to a separate satellite image.  A breakout zoom window is added for images that collected spectra of multiple crop types.}
\label{CropLocations}
\end{center}
\vskip -0.2in
\end{figure}

In this paper we will be working hyperspectral pixel data collected using the NASA Hyperion satellite~\cite{Middleton2013} and organized and meticulously labeled by the USGS.  This data, available online from the USGS as the Global Hyperspectral Imaging Spectral-library of Agricultural crops for Conterminous United States (GHISACONUS)~\cite{GHISACONUS}, is a library of 6,988 spectra, each of which is labeled as one of the five major agricultural crops (e.g., winter wheat, rice, corn, soybeans, and cotton) collected between 2008 and 2015.  The locations for the spectra in the GHISACONUS library are shown in Figure~\ref{CropLocations}.  The spectra seem to be individually labeled pixels from seven different Hyperion images, each of which is apparent as a rectangular shape acquired along the satellite orbit.

\begin{figure}[ht]
\vskip 0.2in
\begin{center}
\centerline{\includegraphics[width=\columnwidth]{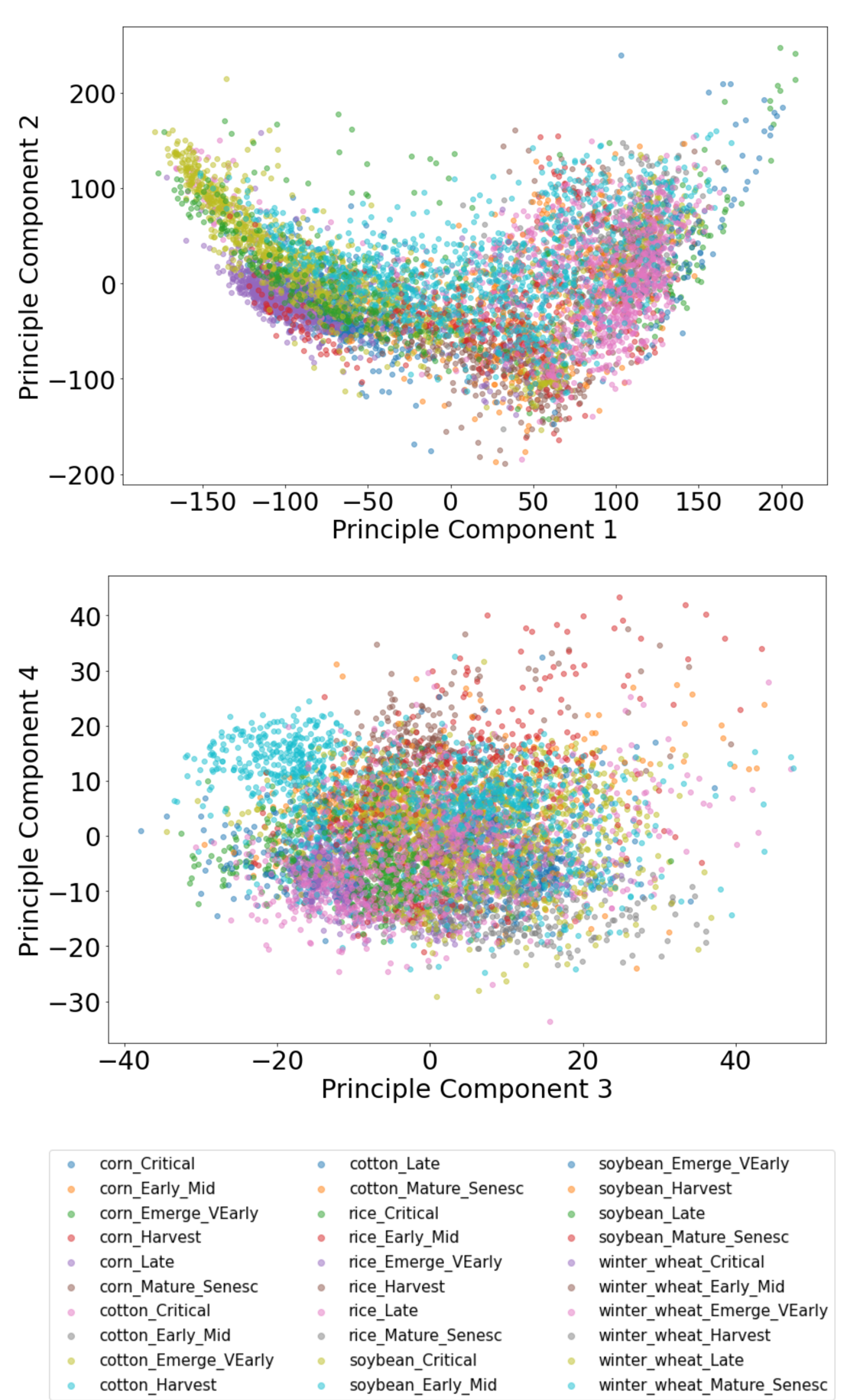}}
\caption{A scattplot of the crop data spectra on the frist two principle components (above) and the third and fourth principle components (below).}
\label{PCA}
\end{center}
\vskip -0.2in
\end{figure}

Each spectrum has 131 wavelengths ranging from 437nm to 2345nm.  Associated with each spectrum is the latitude, longitude, the agroecological zone (AEZ) of the United States for the location (using the United States Department of Agriculture (USDA) Cropland Data Layer (CDL) as reference data).  Also a designation from six different growth stages (emergence/very early vegetative (Emerge VEarly), early and mid vegetative (Early Mid), late vegetative (Late), critical, maturing/senescence (Mature Senesc), and harvest) is included for each crop. The crop growth stage data were derived using crop calendars generated by the Center for Sustainability and the Global Environment (SAGE), University of Wisconsin-Madison.  A scatterplot of the data using the first 4 principle components is shown in Figure~\ref{PCA}.  These PCs contain
67.6, 28.6, 1.3, 0.9 percent of the variance, respectively.

\begin{figure}[ht]
\vskip 0.2in
\begin{center}
\centerline{\includegraphics[width=\columnwidth]{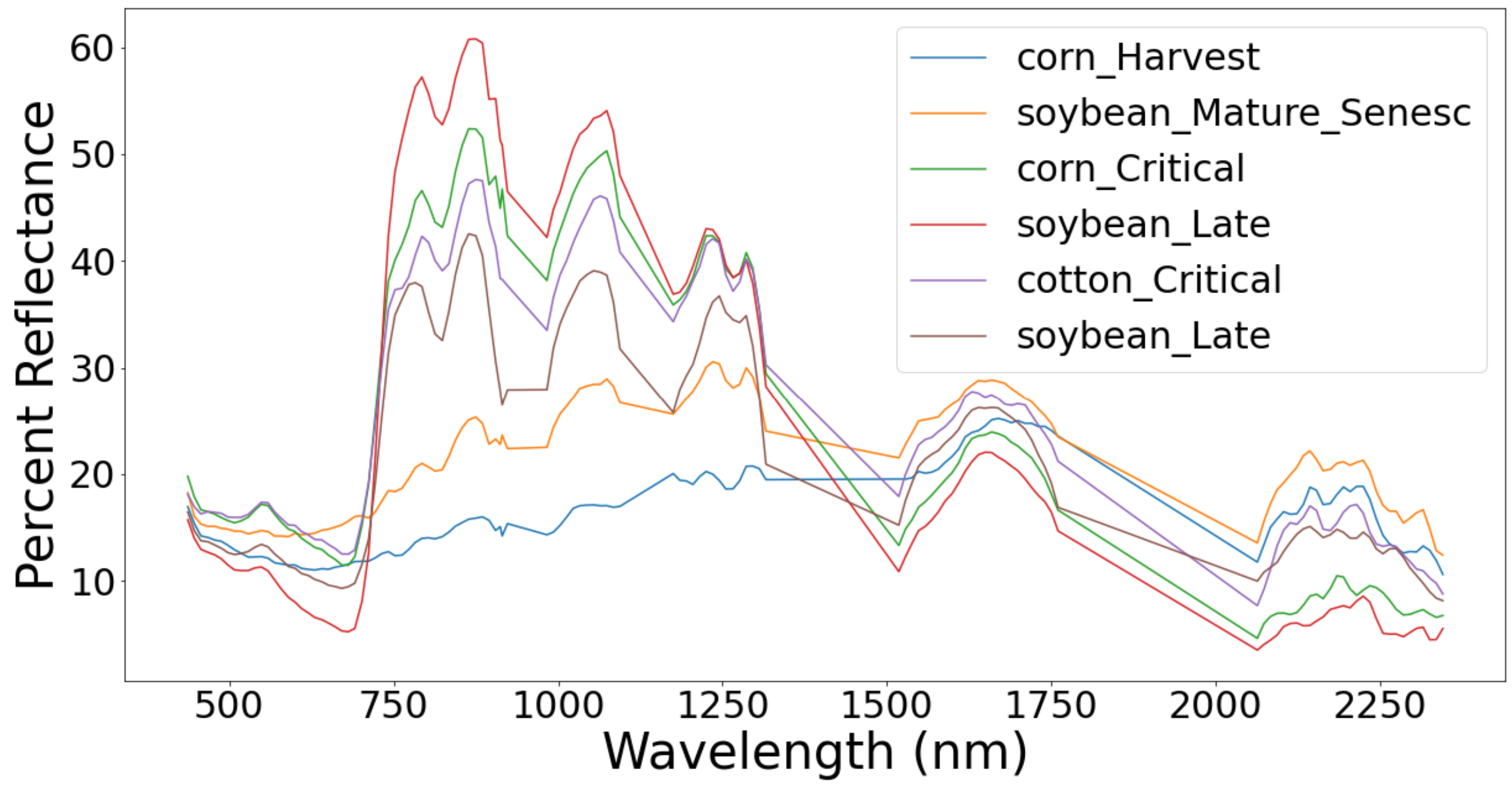}}
\caption{A plot of six spectra from the GHISACONUS spectral library with crop type and growth stage for each spectrum.}
\label{SampleSpectra}
\end{center}
\vskip -0.2in
\end{figure}

\begin{figure}[ht]
\vskip 0.2in
\begin{center}
\centerline{\includegraphics[width=\columnwidth]{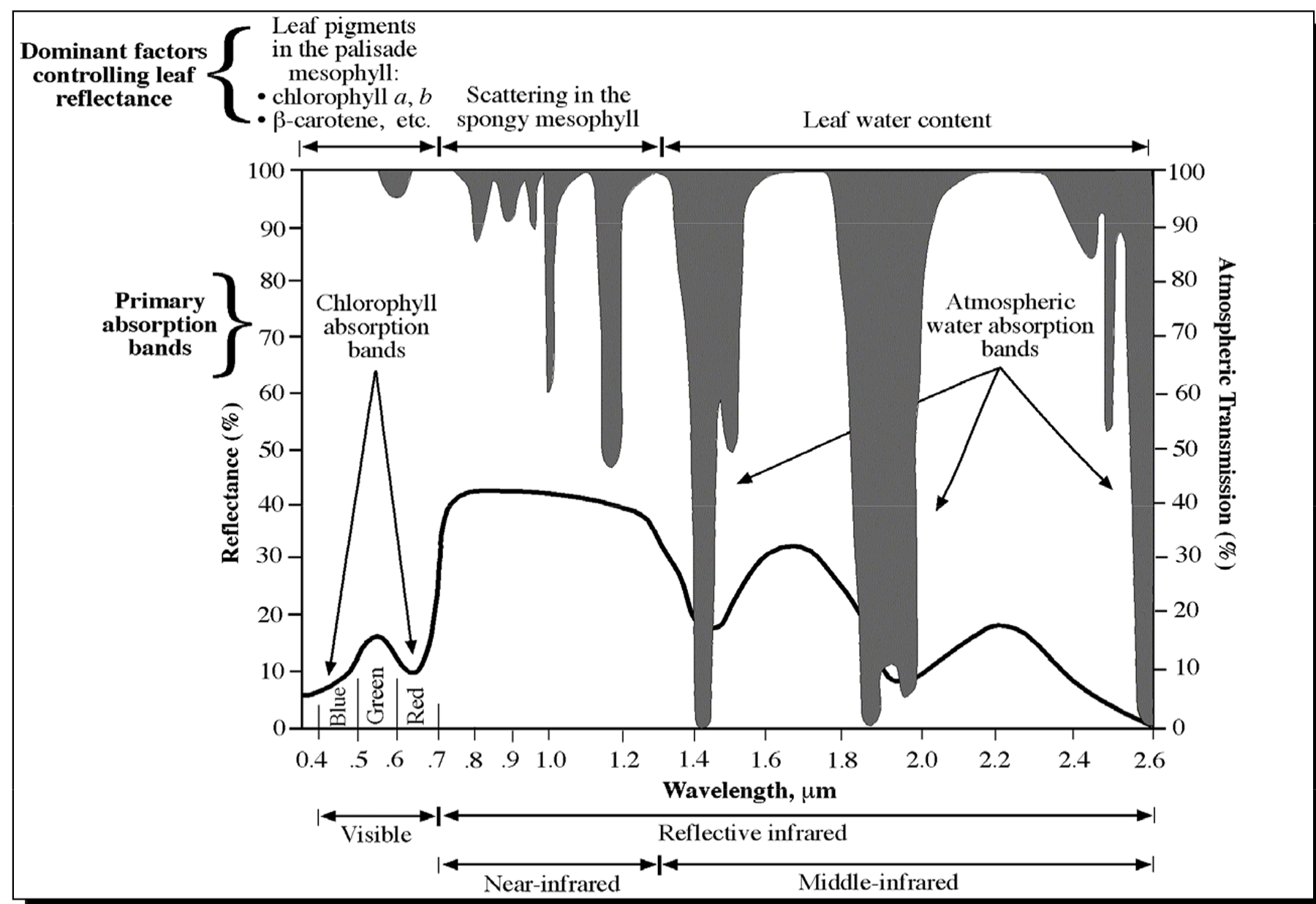}}
\caption{The primary sources of absorbance features in vegetation spectral. (from~\cite{jensen2009remote}.)}
\label{SpectralConstituents}
\end{center}
\vskip -0.2in
\end{figure}

A plot of six spectra from the library is shown in Figure~\ref{SampleSpectra}.  A primary feature associated with vegetation type/growth/health is the sharp rise in reflectance around 750nm which is associated with chlorophyll.  This sharp rise is called the 'red edge' because it occurs just above the wavelengths for visible red light.  Note that the spectra with a strong red edge are in similar growth stages and have an associated peak around 550nm which is visible as the green color of chlorophyll.  Moisture in the vegetation (and possibly atmosphere) absorbs light around 900nm, 1000nm, 1100nm, 1300-1400nm, and 1800-2000nm creating 'valley' features which may be indicators of vegetation moisture level.  Some sources of primary spectral features in vegetation are shown in Figure~\ref{SpectralConstituents} and details can be found in~\cite{jensen2009remote,Hadi2015}.

\section{Methods}
The data from the USGS is in a CSV file.  To facilitate process and open sharing, all code for reading the CSV and processing used for this paper is made available as a Python notebook on Kaggle~\footnote{Available at https://www.kaggle.com/billbasener/ghisaconus-hyperspectral-crop-classification}.

Our primary strategy is to investigate how the crop type and stage information can be leveraged using Bayesian methods to improve classification accuracy.  Therefore, we used Linear Discriminant Analysis (LDA) and Quadratic Discriminant Analysis (QDA) from scikit-learn leveragin both the class prediction and class posterior probability methods.  We focused on these Gaussian model methods because hyperspectral classes for a single material is generally believed to be well-approximated by a Gaussian or similarly contoured distribution~\cite{ManolakisMardenKerekesShaw2001, MardenManolakis2003, Mano03}.  However, the variation in spectra over the crop growth stages implies that the distribution per crop-stage pair may be Gaussian, the distribution for a given crop over all stages is not expected to be Gaussian.  Thus, Bayesian methods~\cite{GhoshDelampadySamanta2006} allow us to compute probabilities for each crop-stage pair and then combine probabilities across stages of growth to obtain probabilities per crop type.

The algorithms we test are: \textbf{LDA:} LDA using the predict method using just the crop type as prediction variable.  \textbf{LDA Bayes(MMP):} LDA using the predict probability method with the combined crop type and growth stage as the prediction variable to get a joint probability distribution for type and state; and then computing the marginal probability for each crop type by summing probabilities for all growth stages for the type and predicting the crop type with the maximal marginal probability (MMP) as the predicted class. \textbf{LDA Bayes(MJP):}LDA using the predict probability method with the combined crop type and growth stage as the prediction variable to get a joint probability distribution for type and state; and then using the crop type from the copr-stage pair with the highest joint probability as the predicted type.  \textbf{QDA}, \textbf{QDA Bayes(MMP)}, and \textbf{QDA Bayes(MJP)} are analogous using QDA in place of LDA, computing a seperate covariance for each class.  Because the number of features is 131 (number of wavelengths), some classes are small compared to the number of features and so we used a regularization parameter (\texttt{reg\_param} in scikit-learn) in QDA.  This parameter was optimized by choosing the best results among the values $[0.01, 0.05, 0.1, 0.2, ..., 0.9, 1]$.  We also tested two neural networks.  One NN has a single hidden layer with 256 ReLu neurons, 5\% dropout dropout layer, training for 100 epochs.  The second differs by having two hidden layers of 256 ReLu neurons, each followed by a 5\% dropout layer.  We included these neural networks as reasonable nonlinear NNs for comparison, but do not claim that they highly optimized.  We assume constant priors in all Bayesian classifiers.  Accuracy would likely improve if we computed priors either from class frequency or spatial locations, but we chose to use noninformative priors to focus on the distributions of the spectra.

To illustrate the the Bayesian approach, Tables~\ref{BayesQDA0} through~\ref{BayesQDA2} show the joint probability distribution over crop type and stage for three different spectra.  In the Max Marginal Probability output, the marginal probability for each crop is computed by summing down the columns and this is used for classification.  We also compute the result from a Max Joint Probability where the crop for the crop-stage pair with the max probability is used for classification.  The Max Marginal Probability method is more rigorous in the Bayesian sense of computing a marginal probability for the crop.

\begin{table}[t]
\caption{Joint probability for Corn in the Critical stage using QDA Bayes.  Note that in this example, the probability is distributed across two different crops at the same stage, but the highest probability is assigned to the correct crop.  The probability assigned to the correct crop-stage pair is indicated in bold.}
\label{BayesQDA1}
\vskip 0.15in
\begin{center}
\begin{small}
\begin{sc}
\begin{tabular}{lccccc}
\toprule
                & Corn  & Cotton    & Rice  & Soybn   & Wheat \\
Stage           &       &           &       &           & \\
\midrule
Critical        & \textbf{0.58}  & 0.0    & 0.0  & 0.42   & 0.0 \\
Erl\_Mid       & 0.0  & 0.0    & 0.0  & 0.0   & 0.0 \\
Emg\_VE   & 0.0  & 0.0    & 0.0  & 0.0   & 0.0 \\
Harvest         & 0.0  & 0.0    & 0.0  & 0.0   & 0.0 \\
Late            & 0.0  & 0.0    & 0.0  & 0.0   & 0.0 \\
Mat\_Sen   & 0.0  & 0.0    & 0.0  & 0.0   & 0.0 \\
\bottomrule
\end{tabular}
\end{sc}
\end{small}
\end{center}
\vskip -0.1in
\end{table}
\begin{table}[t]
\caption{Joint probability for a spectrum Corn in the Mature\_Senesc stage using QDA Bayes.  Note that in this example, the probability is distributed across the correct crop at two different stages.  The probability assigned to the correct crop-stage pair is indicated in bold.}
\label{BayesQDA2}
\vskip 0.15in
\begin{center}
\begin{small}
\begin{sc}
\begin{tabular}{lccccc}
\toprule
                & Corn  & Cotton    & Rice  & Soybn   & Wheat \\
Stage           &       &           &       &           & \\
\midrule
Critical        & 0.48  & 0.0    & 0.0  & 0.0   & 0.0 \\
Erl\_Mid       & 0.0  & 0.0    & 0.0  & 0.0   & 0.0 \\
Emg\_VE   & 0.0  & 0.0    & 0.0  & 0.0   & 0.0 \\
Harvest         & 0.0  & 0.0    & 0.0  & 0.0   & 0.0 \\
Late            & 0.0  & 0.0    & 0.0  & 0.0   & 0.0 \\
Mat\_Sen   & \textbf{0.52}  & 0.0    & 0.0  & 0.0   & 0.0 \\
\bottomrule
\end{tabular}
\end{sc}
\end{small}
\end{center}
\vskip -0.1in
\end{table}
\begin{table}[t]
\caption{Joint probability for a spectrum of Winter Wheat in the Mature\_Senesc stage using QDA Bayes.  Note that in this example, the probability is distributed across three crop-stage pairs but the correct crop is given the second highest probability so this is an incorrect classification.  The probability assigned to the correct crop-stage pair is indicated in bold.}
\label{BayesQDA2}
\vskip 0.15in
\begin{center}
\begin{small}
\begin{sc}
\begin{tabular}{lccccc}
\toprule
                & Corn  & Cotton    & Rice  & Soybn   & Wheat \\
Stage           &       &           &       &           & \\
\midrule
Critical        & 0.23  & 0.0    & 0.0  & 0.44   & 0.0 \\
Erl\_Mid       & 0.0  & 0.0    & 0.0  & 0.0   & 0.0 \\
Emg\_VE   & 0.0  & 0.0    & 0.0  & 0.0   & 0.0 \\
Harvest         & 0.0  & 0.0    & 0.0  & 0.0   & 0.0 \\
Late            & 0.00  & 0.0    & 0.0  & 0.0   & 0.0 \\
Mat\_Sen   & 0.0  & 0.0    & 0.0  & 0.0   & \textbf{0.33} \\
\bottomrule
\end{tabular}
\end{sc}
\end{small}
\end{center}
\vskip -0.1in
\end{table}

\section{Results}
The accuracy for each algorithms is shown in Figure~\ref{accuracies}. Each algorithm was tested with 10-fold cross validation.  The accuracy shown is mean accuracy over the folds and the standard deviation used for the confidence interval provided is the standard deviation of the accuracies over the folds.

\begin{table}[t]
\caption{Classification accuracies.  Along with each accuracy is the 95\% confidence interval from 10 fold cross validation.  The abbreviations are MMP = Max Marginal Probability, MJP = Max Joint Probability, RP = regularization parameter, HL = Hidden Layer.}
\label{accuracies}
\vskip 0.15in
\begin{center}
\begin{small}
\begin{sc}
\begin{tabular}{lcccr}
\toprule
Algorithm & Accuracy$\pm 2\sigma$ &  Best \\
\midrule
LDA                             & 77.4$\pm$ 2.8&  $$ \\
LDA Bayes (MMP)                 & 83.8$\pm$ 2.2&  $$ \\
LDA Bayes (MJP)                 & 83.7$\pm$ 2.4&  $$\\
QDA (RP=0.01)                   & 82.4$\pm$ 2.4&          \\
QDA Bayes (MMP, RP=0.5)         & 85.3$\pm$ 2.9&  $\surd$ \\
QDA Bayes (MJP, RP=0.5)         & 85.3$\pm$ 2.9&  $\surd$\\
Neural Network (1HL)            & 76.0$\pm$ 3.2&          \\
Neural Network (2HL)            & 80.7$\pm$ 3.7&          \\
\bottomrule
\end{tabular}
\end{sc}
\end{small}
\end{center}
\vskip -0.1in
\end{table}

\begin{figure}[ht]
\vskip 0.2in
\begin{center}
\centerline{\includegraphics[width=\columnwidth]{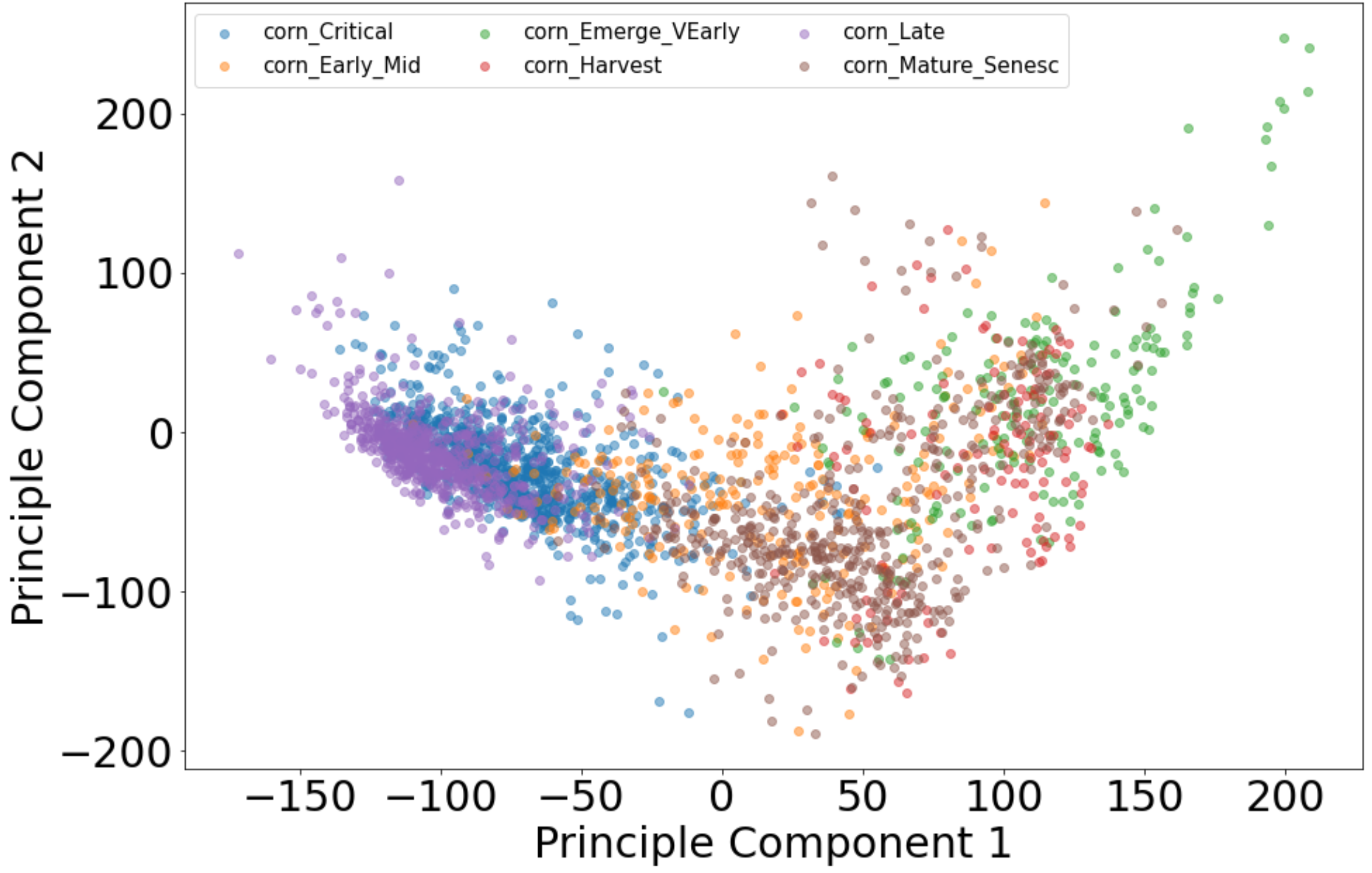}}
\caption{Locations of the crops associated with the collected spectra, with each spectrum indicated by a marker colored by the crop type.  Observe that the collections come in rectangular regions, each of which corresponds to a separate satellite image.  A breakout zoom window is added for images that collected spectra of multiple crop types.}
\label{PCA_Corn}
\end{center}
\vskip -0.2in
\end{figure}

\begin{figure}[ht]
\vskip 0.2in
\begin{center}
\centerline{\includegraphics[width=\columnwidth]{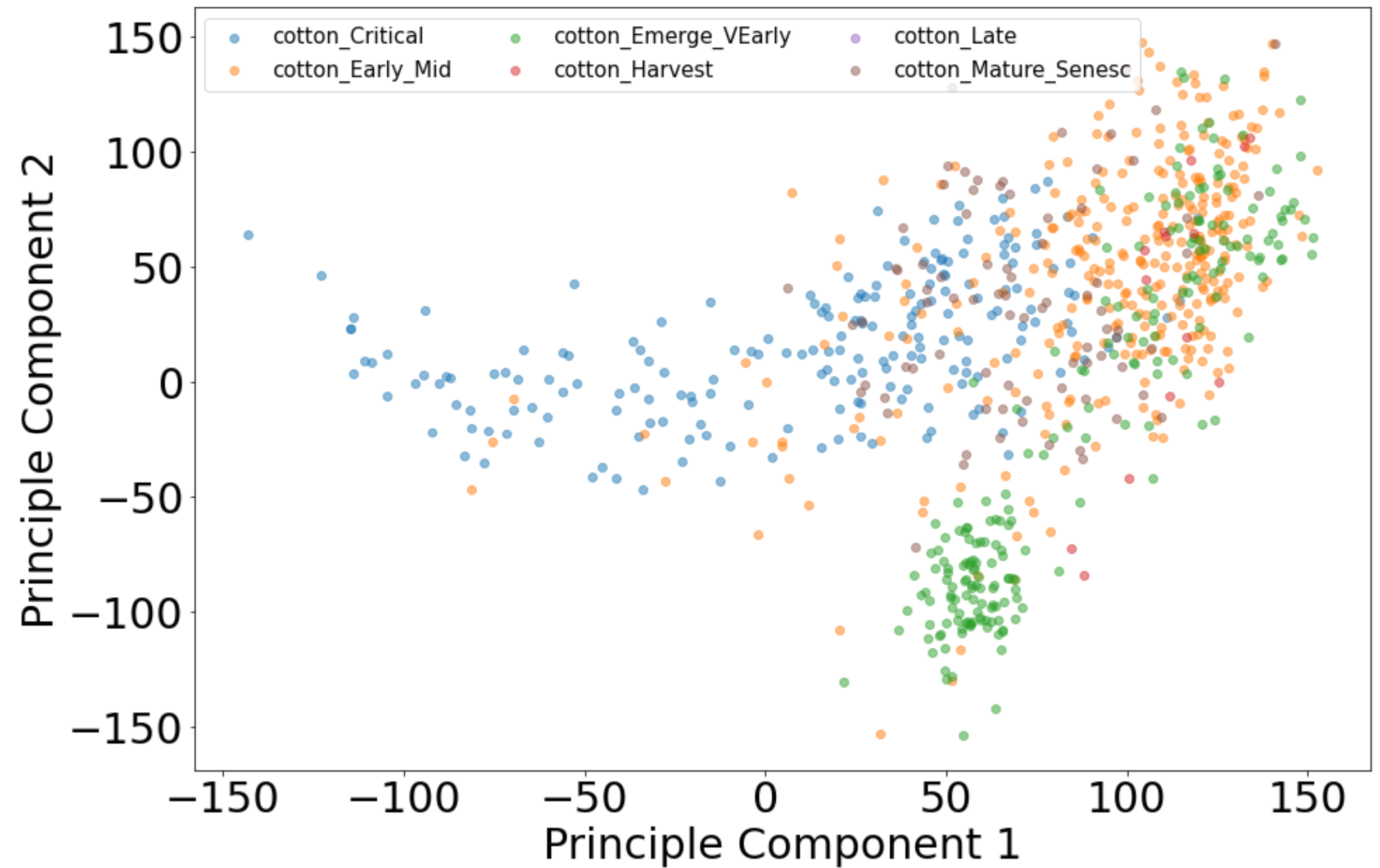}}
\caption{Scatterplot of the cotton spectra on the frist two principle components computed from the full data, colored by growth stage}
\label{PCA_Cotton}
\end{center}
\vskip -0.2in
\end{figure}

\begin{figure}[ht]
\vskip 0.2in
\begin{center}
\centerline{\includegraphics[width=\columnwidth]{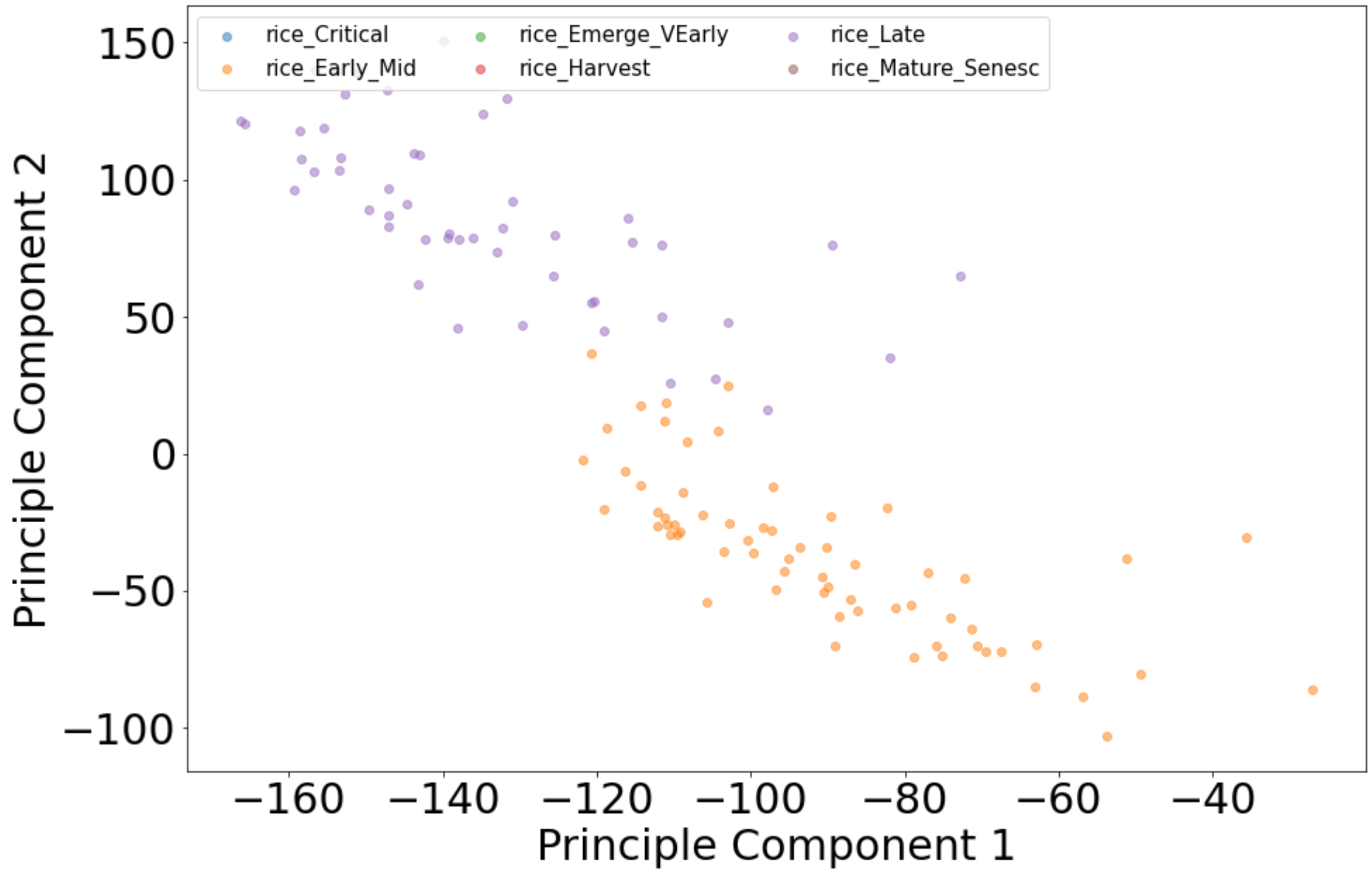}}
\caption{Scatterplot of the rice spectra on the frist two principle components computed from the full data, colored by growth stage}
\label{PCA_Rice}
\end{center}
\vskip -0.2in
\end{figure}

\begin{figure}[ht]
\vskip 0.2in
\begin{center}
\centerline{\includegraphics[width=\columnwidth]{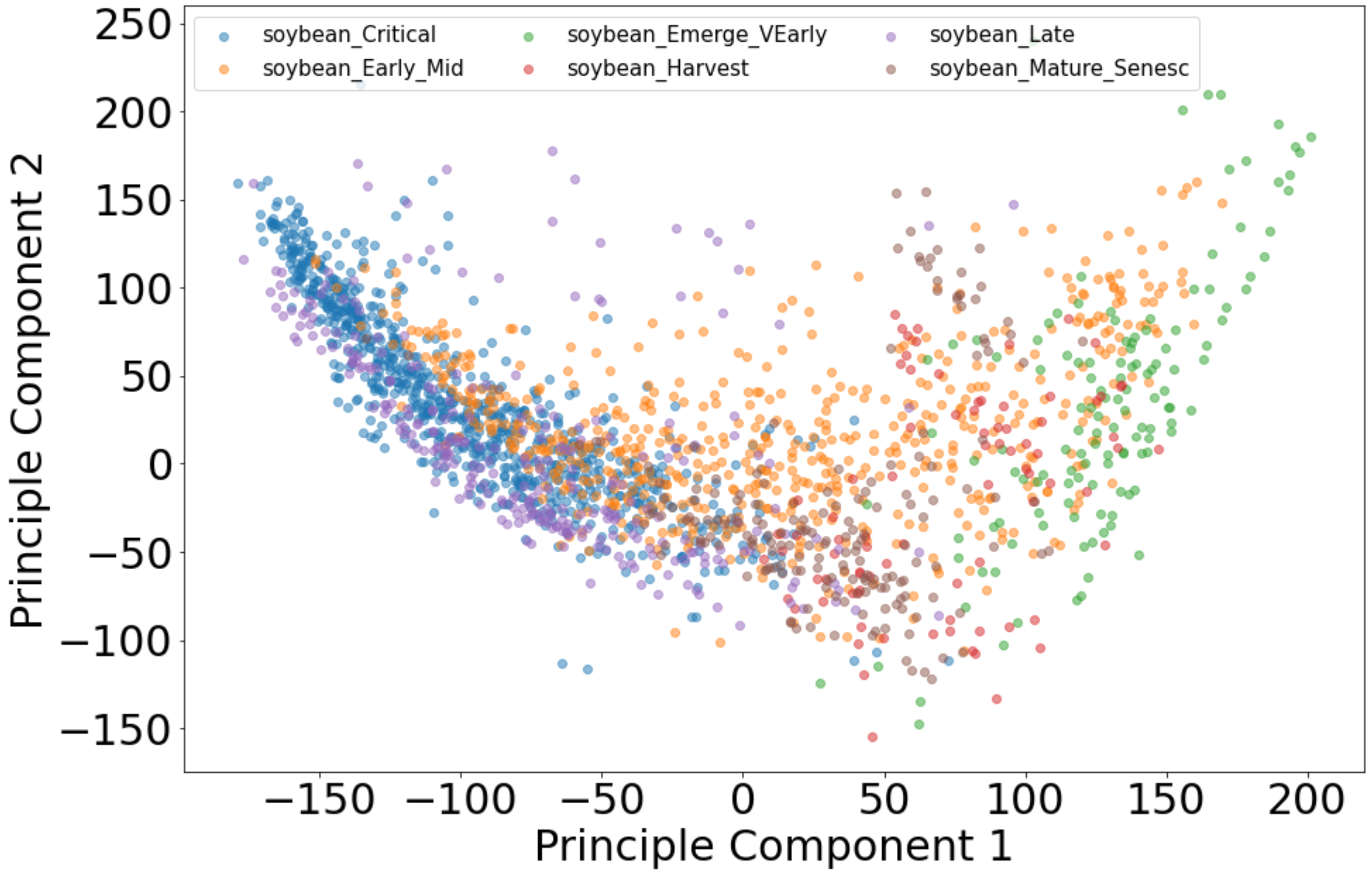}}
\caption{Scatterplot of the soybean spectra on the frist two principle components computed from the full data, colored by growth stage}
\label{PCA_Soybean}
\end{center}
\vskip -0.2in
\end{figure}

\begin{figure}[ht]
\vskip 0.2in
\begin{center}
\centerline{\includegraphics[width=\columnwidth]{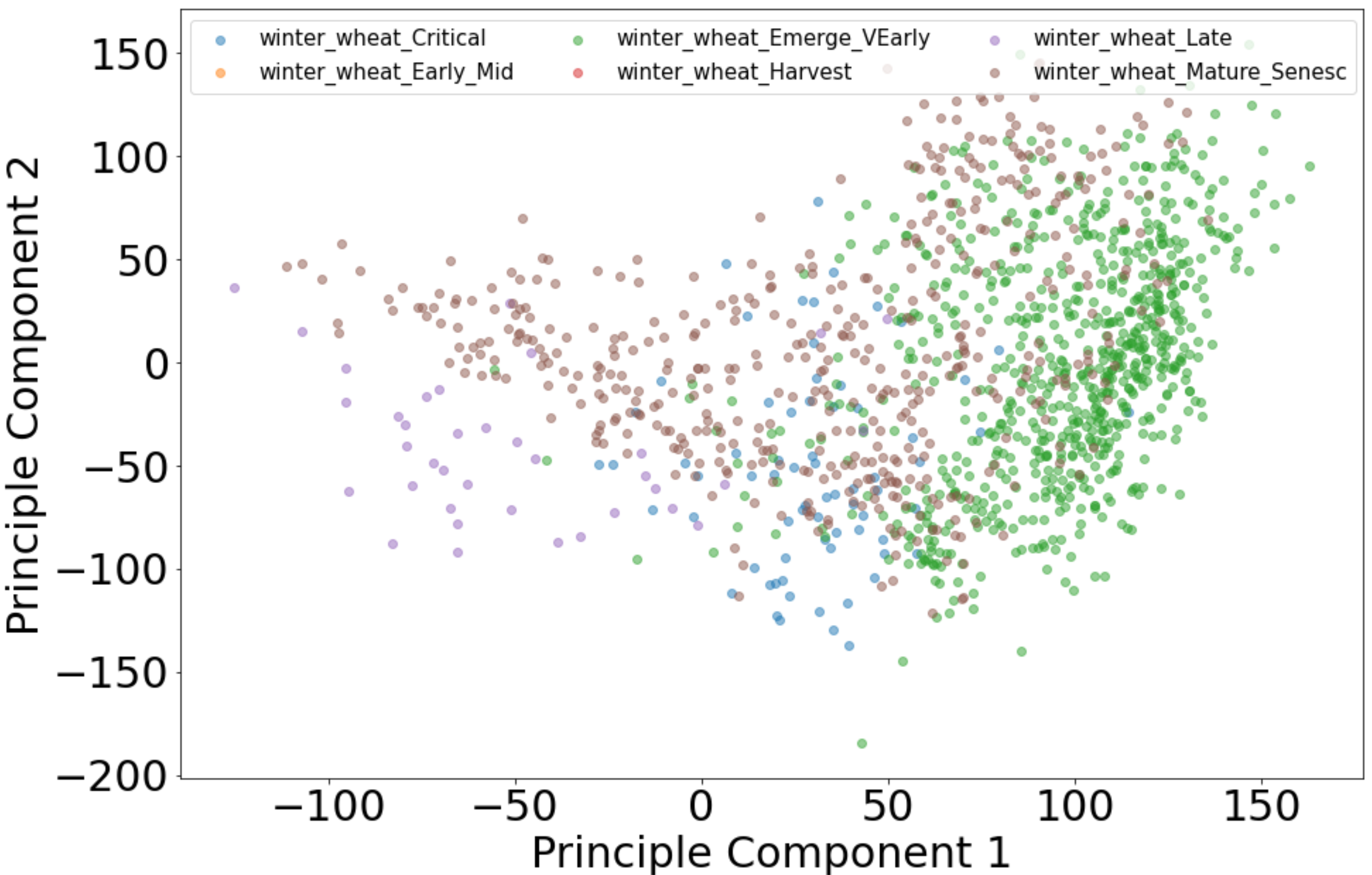}}
\caption{Scatterplot of the wheat spectra on the frist two principle components computed from the full data, colored by growth stage}
\label{PCA_Wheat}
\end{center}
\vskip -0.2in
\end{figure}

To visualize the data, we computed the first two principle components for the dataset as a whole.  Principle component 1 contains 67\% of the variance while principle component 2 contains 28\% of the variance, so these plots give a reasonable representation of the data containing about 95\% of the variance.  This suggests the vegetation spectra have an inherent dimension not much more than 2, which is less than the 10-20 dimensions estimated for full images with more material variation~\cite{BasenerSchlammMessinger2008}.  The projection of the data for each crop, colored by growth stage, and shown in Figures~\ref{PCA_Corn} through~\ref{PCA_Wheat}.  Because the Very Early and Senescenced growth stages tend to appear toward the right-hand side of these plots while Critical and Late tend toward the left, it appears in each case that principle component 1 is in the direction of increased chlorophyll amount as measured for example by the red edge.

\section{Conclusion}

The first conclusion is that clearly the crop classes do not have equal Gaussian distribution as the LDA classifier had lower accuracy than the more complex models; with for example QDA performed well above LDA relative to the confidence intervals.  Furthermore, the Bayesian QDA methods performed the best, suggesting that the crop-stage classes in the joint probability distribution do not have equal Gaussian distributions.  However, the high variance nonlinear NN with two 256 neuron layers did not outperform the Bayesian QDA, suggesting that the assumption that each crop-growth stage class can reasonably be approximated by a Gaussian is at least reasonable.  All of these observations are supported by the scatterplots shown in Figures~\ref{PCA_Corn} through~\ref{PCA_Wheat}.

It is interesting that for both LDA and QDA, the Bayesian MMP and MJP methods performed very similarly.  Also, because the QDA methods required regularization, it seems likely that more training data would further improve accuracy.

\bibliography{my_refs2}

\begin{thebibliography}{10}

\bibitem{Olsen_Bergman_Resmini_1997}
R.~Olsen, S.~Bergman, and R.~Resmini,
\newblock ``Target detection in a forest environment using spectral imagery,''
\newblock {\em processamento de imagensprincipais componentes}, pp. 46--56,
  1997.

\bibitem{SchottBook2007}
J.~Schott,
\newblock {\em Remote Sensing: The Imaging Chain Approach},
\newblock Oxford University Press, 2nd edition, 2007.

\bibitem{Middleton2013}
Elizabeth Middleton, Stephen Ungar, Daniel Mandl, Lawrence Ong, Stuart Frye,
  Petya Campbell, David Landis, Joseph Young, and Nathan Pollack,
\newblock ``The earth observing one (eo-1) satellite mission: Over a decade in
  space,''
\newblock {\em IEEE JOURNAL OF SELECTED TOPICS IN APPLIED EARTH OBSERVATIONS
  AND REMOTE SENSING}, vol. 6, pp. 243, 04 2013.

\bibitem{GHISACONUS}
I~Thenkabail, P.and~Aneece,
\newblock ``Global hyperspectral imaging spectral-library of agricultural crops
  for conterminous united states v001,'' 2019,
\newblock NASA EOSDIS Land Processes DAAC. Accessed 2022-01-18,
  \url{https://doi.org/10.5067/Community/GHISA/GHISACONUS.001}.

\bibitem{jensen2009remote}
John~R Jensen,
\newblock {\em Remote sensing of the environment: An earth resource perspective
  2/e},
\newblock Pearson Education India, 2009.

\bibitem{Hadi2015}
Hadi Hadi,
\newblock {\em Multivariate statistical analysis for estimating grassland leaf
  area index and chlorophyll content using hyperspectral data},
\newblock Ph.D. thesis, 06 2015.

\bibitem{ManolakisMardenKerekesShaw2001}
J.~Kerekes D.~Manolakis, D.~Marden and G.~Shaw,
\newblock ``Statistics of hyperspectral imaging data,''
\newblock 2001, vol. 4381 of {\em Proc. SPIE}, pp. 308--316.

\bibitem{MardenManolakis2003}
D.~Marden and D.~Manolakis,
\newblock ``Modeling hyperspectral imaging data,''
\newblock 2003, vol. 5093 of {\em Proc. SPIE}.

\bibitem{Mano03}
D.~Marden and D.~Manolakis,
\newblock ``Modeling hyperspectral imaging data using elliptical contoured
  distributions,''
\newblock 2003, vol. 5093 of {\em Proc. SPIE}.

\bibitem{GhoshDelampadySamanta2006}
M.~Delampady J.~Ghosh and T.~Samanta,
\newblock {\em An Introduction to Bayesian Analysis: Theory and Methods},
\newblock Springer, 2006.

\bibitem{BasenerSchlammMessinger2008}
Ariel Schlamm, David Messinger, and William Basener,
\newblock ``Geometric estimation of the inherent dimensionality of a single
  material cluster in multi- and hyperspectral imagery,'' 2008.

\end{thebibliography}
\bibliographystyle{IEEEbib}

\end{document}